\documentclass[letterpaper, 10 pt, conference]{template/ieeeconf} 
\IEEEoverridecommandlockouts                              
\usepackage{bm}
\usepackage{algorithm}
\usepackage{algpseudocode}
\usepackage{amsmath}
\usepackage{graphicx} 
\usepackage{subfig} 
\usepackage{multirow}
\usepackage{nicefrac}
\usepackage{dblfloatfix}  
\usepackage[export]{adjustbox}  
\usepackage{hyperref}
\usepackage{comment}
\usepackage{amsfonts} % added by Rans

\DeclareMathOperator*{\argmin}{argmin}

\newcommand{\x}{\textbf{x}}

\newcommand{\m}{\bm{\mu}}
\newcommand{\bS}{\bm{\Sigma}}
\newcommand{\bP}{\textbf{P}}
\newcommand{\bR}{\textbf{R}}
\newcommand{\bt}{\textbf{t}}

\title{\LARGE \bf
	Dynamic Hilbert Maps: \\ Real-Time Occupancy Predictions in Changing Environments
}

\author{Vitor Guizilini$^{1,2}$, Ransalu Senanayake$^{1}$, and Fabio Ramos$^{1,3}$ % 
	\thanks{The authors are with the $^1$School of Computer Science, at The University of Sydney, Australia, $^2$Toyota Research Institute, USA, and $^3$NVIDIA Research, USA. Emails: {\tt\small \{ vitor.guizilini;ransalu.senanayake;fabio.ramos}
    {\tt\small \}@sydney.edu.au} }%
}

\begin{document}
	
\maketitle
\thispagestyle{empty}
\pagestyle{empty}

%%%%%%%%%%%%%%%%%%%%%%%%%%%%%%%%%%%%%%%%%%%%%%%%%%%%%%%%%%%%%%%%%%%%%%%%%%%%%%%%

\begin{abstract}
	
This paper addresses the problem of learning instantaneous occupancy levels of dynamic environments and predicting future occupancy levels. Due to the complexity of most real-world environments, such as urban streets or crowded areas, the efficient and robust incorporation of temporal dependencies into otherwise static occupancy models remains a challenge. We propose a method to capture the spatial uncertainty of moving objects and incorporate this uncertainty information into a continuous occupancy map represented in a rich high-dimensional feature space. Experiments performed using LIDAR data verified the real-time performance of the algorithm.
% * <ransalu.s@gmail.com> 2018-02-18T12:51:32.287Z:
% 
% > Such model 
% a model or models?
% 
% ^. Such models should be able to propagate the location of mapped objects into the future, accurately predicting their position and thus allowing for better path planning and obstacle avoidance routines.
	
\end{abstract}

\section{INTRODUCTION}
\label{sec:intro}

Autonomous vehicles are no longer restricted to controlled test environments and have begun their transition to unstructured real-world environments. In order to operate a vehicle autonomously, its control algorithms require a representation of the surroundings. For autonomous navigation, this representation usually takes the form of an occupancy map, describing which areas are empty (safe for traversal) and which areas are occupied (would result in a collision).

The straightforward approaches to static occupancy mapping rely on a grid-based non-overlapping discretization of the environment \cite{Elf1989}. Because grid cells are updated individually without considering the relationship among cells, this discretization process completely discards spatial or spatiotemporal dependencies. Furthermore, the discretized representation quickly becomes infeasible for larger datasets, especially when dealing with volumetric data. The Hilbert Mapping (HM) framework \cite{RamOtt2015,DohWanEng,DohWanEng2016IROS} is an alternative to grid maps and can produce a continuous representation of occupancy states in a much lower computational cost.

\begin{figure}[b]
	\vspace{-0.4cm}
	\centering
    \subfloat[Dataset at $t=0$]{\includegraphics[width=0.14\textwidth]{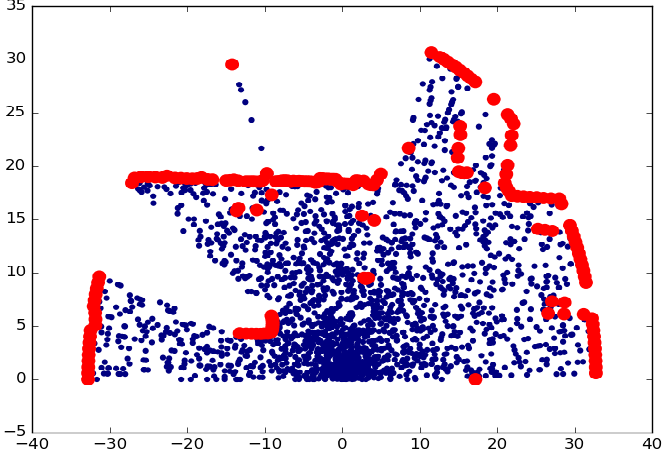}} 
	\subfloat[Future occupancy prediction]{\includegraphics[width=0.3\textwidth]{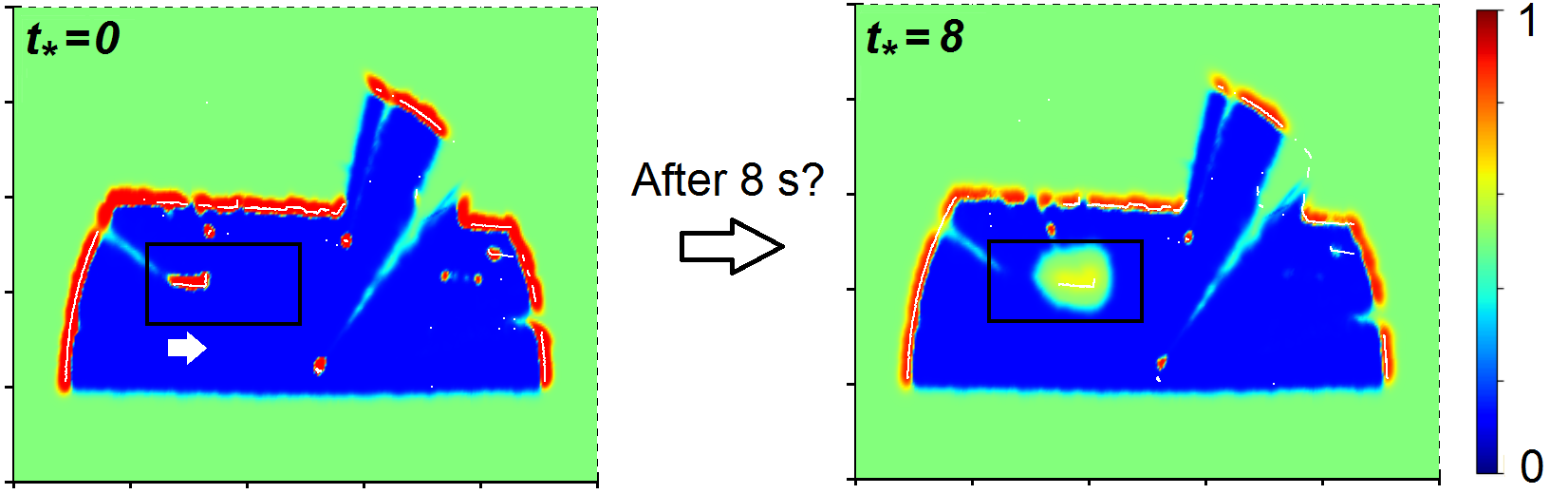}} 
	\caption{\small (a) Data-frame captured by a LiDAR (blue: laser beams and red: laser hit points). The vehicle inside the rectangle moves from left to right (b) The current and future (8 seconds) occupancy maps produced by the Dynamic Hilbert maps (DHM) algorithm. Indicating the uncertainty of future predictions, the occupancy probability (red indicates highly probable) of the position of the vehicle and its surrounding is relatively low. The future prediction is represented as a spatial distribution peaked at one point which drops down radially, making such a map ideal for safer path planning.}	
	\label{fig:moti}
\end{figure}

%\subsection{Related Work}

Typically, the objective of occupancy representation is to build a map that can later be used for off-line path planning. Nevertheless, with the requirement of operating robots in real-world environments, it is essential to take dynamics of the environment into account and adjust control policies accordingly. Dynamic occupancy mapping can be categorized into three classes: 1) building static occupancy maps in the presence of dynamic objects, 2) mapping the long-term dynamics of the environment, and 3) mapping the short-term dynamics of the environment. Most of the early research studies on dynamic occupancy mapping fall under the first category in which the dynamic objects are treated as spurious data and remove them to build a robust static map \cite{MeyBeiBug2012AAAI,HahTri2003ICRA,StaBug2005AAAI,NusReu2016arXiv}. In the second category, an occupancy pattern is obtained over a long period \cite{MeyBeiBug2012AAAI, KraFenCieDonDuc2014ICRA, SenOccRam2017ICRA, SenRam2017CoRL} and such a map can later be used for global path planning. This paper focuses on the third category---mapping short-term dynamics. The short-term dynamics are important not only for understanding the instantaneous changes in the environment but also for making predictions into the future. 

% Historically, most existing occupancy mapping techniques are only able to deal with static environments \cite{ThrBurFox2005}, in which only the vehicle itself is moving. However, this is unrealistic for most real-world problems, in which objects change their position over time without any predetermined behaviors. A common solution is the introduction of parallel processes for dynamic object tracking \cite{BreGinDil2010,MeyBeiBur2014}, but while computationally efficient this approach discards crucial spatio-temporal dependencies between data observed in different timesteps. The incorporation of these dependencies directly into the occupancy model would make it possible to predict future environment states (i.e. areas more likely to be occupied given currently observed motion), and take this information into account during the decision-making process.  

%Even though there have been attempts to adapt grid-based occupancy models to dynamic environments \cite{BreGinDil2010,TanThoWolBus2014,TipMeyBeiBur2012}, here we focus on continuous models, that are more attractive due to the above mentioned properties and can make predictions in to the future. 
To the best of our knowledge, there are only two main techniques to model short-term occupancy and make occupancy predictions into the future. An extension to GPOM, named dynamic Gaussian process (DGP) maps, was proposed in \cite{CalRam2015} to model the occupancy state of dynamic environments. It proposes a novel covariance function that captures space-time statistical dependencies, thus enabling predictions on future occupancy states. The cubic computational complexity of Gaussian process (GP) classification \cite{RasWil2005} makes DGPs infeasible in real-time dynamic occupancy mapping. As an alternative, a spatiotemporal extension to HMs  (STHMs) was proposed in \cite{SenOttCalRam2016}. This framework uses hinged features to combine temporal variability into the spatial domain modeled by an underlying motion modeling framework.

This paper proposes a novel methodology for spatiotemporal occupancy modeling that builds upon the HM framework, generalizing it to dynamic environments. As shown in Figure~\ref{fig:moti}, the proposed approach can make predictions into the future. Note that the area of uncertainty around the moving vehicle is much larger, due to the increase in uncertainty, which is useful for safely executing motion plans. However, unlike \cite{SenOttCalRam2016}, where these dependencies are modeled by a hinged kernel that receives predicted spatial coordinates from a Gaussian process \cite{RasWil2005}, the proposed methodology works by directly updating the feature vector that projects input points into the reproducing kernel Hilbert space (RKHS) \cite{SchSmo2001} for classification, changing its shape to accommodate external motion. A probabilistic dynamic model is incrementally learned for each observed object, using point-cloud alignment techniques on clustered data, and localization uncertainties are also propagated to the occupancy model, accounting for sensor inaccuracies and accumulated drift. The result is an efficient framework capable of tracking multiple 3D objects in real-time, while accurately using this information to probabilistically predict the occupancy state of the environment at arbitrary spatio-temporal resolutions.

The proposed Dynamic Hilbert Maps (DHM) framework also shares similarities with the broader field of dynamic object tracking \cite{Wan2014,HelLevThrSav2014,DewCasTipBur2016}, in the sense that it detects and segments external motion over time.  However, while these methods are limited to point-cloud modeling and tracking, ours directly incorporates this information into an incremental occupancy map, which can then be propagated to future or past timesteps to predict the environment at arbitrary resolutions. Additionally, while most dynamic object tracking approaches rely on an initial supervised training stage, either as background filters \cite{UshWolWalEus2017}, rigidly attached sample points \cite{WanPosNew2015} or deep recurrent neural networks \cite{DeqAl2017}, ours is able to build large-scale 3D probabilistic predictive occupancy models without the prior knowledge of similar environments. Although our primary objective is not object tracking, positions and shapes of objects can be easily extracted from DHMs, making DHMs a generalization of object tracking.

\section{Static Hilbert Maps}
\label{sec:hm}

Following LARD \cite{RamOtt2015, GuiRam2016}, we define a collection of hinged locations $\tilde{\mathcal{X}}$ that act as inducing points \cite{SneGha2006}. With analogy to a multivariate Gaussian shape, these hinged locations have a center $\bm{\mu} \in \mathbb{R}^3$ alongside another matrix $\bS \in \mathbb{R}^{3 \times 3}$ to denote how far the measurements affect in each direction . With the $M$ hinged locations $\tilde{\mathcal{X}} = \{ \tilde{\mathbf{x}}_m \}_{m=1}^M = \{ (\bm{\mu}_m, \bS_m) \}_{m=1}^M$, the occupancy probability of any point in the environment $ \mathbf{x}_* \in \mathbb{R}^3$ can be computed using a logistic model,
%\vspace{-5pt}
\begin{equation}  \label{eq:estimate} \small
p(y_* = 1 \vert \textbf{x}_*,\textbf{w}, \tilde{\mathcal{X}}) = \big( 1 + \exp \left( - \textbf{w}^\top\Phi(\textbf{x}_*) \right) \big)^{-1},
\end{equation}
with a feature vector defined as:
\begin{align} \label{eq:lard} \small
\Phi(\textbf{x}_*,\tilde{\mathcal{X}} ) &= \left[ 
k(\mathbf{x}_*, \tilde{\mathbf{x}}_1) \hspace{0.05cm} , \hspace{0.05cm} 
k(\mathbf{x}_*, \tilde{\mathbf{x}}_2) \hspace{0.05cm} , \hspace{0.05cm} 
\dots \hspace{0.05cm} , \hspace{0.05cm} 
k(\mathbf{x}_*, \tilde{\mathbf{x}}_M)
\right], \\
k(\x_*,\tilde{\mathbf{x}}_m) &= \exp \left( - \frac{1}{2} ( \x_* - \bm{\mu}_m )^\top\bS_m^{-1} ( \x_* - \bm{\mu}_m ) \right).
\end{align}

Learning the model involves two steps. Firstly, the dataset $\mathcal{D}$ is used to determine the hinge locations $\tilde{\mathcal{X}}$ using a clustering algorithm \cite{GuiRam2016}. Then, as the crucial step,  the parameters $\mathbf{w}$ are learned by minimizing the objective function of the regularized logistic regression,
\begin{equation} \label{eq:rnll}
\small
\sum_{n=1}^N \left( 1 + \exp\left( -y_n\textbf{w}^\top  \Phi(\textbf{x}_n) \right) \right) + \lambda_1 \lVert \textbf{w} \rVert_2^2 + \lambda_2\lVert \textbf{w} \rVert_1,
\end{equation}
where $\lVert \cdot \rVert$ is the norm and $\lambda$ are the regularizations weights used to regularize the classifier. $\Phi(\x_n)$ is computed similar to (\ref{eq:lard}) by evaluating $k(\x_n,\tilde{\mathbf{x}}_m)$ values. Importantly, since the objective function is represented as a sum of data points, stochastic gradient descent (SGD) \cite{Bot2010COMPSTAT} can be used. 

\section{Dynamic Hilbert Maps}
\label{sec:tphm}

This section introduces the proposed methodology for dynamic occupancy modeling, that builds upon the Hilbert Maps framework reviewed in Section~\ref{sec:hm}. %As outlined in Section~\ref{sec:intro},
Our objective is to build short-term occupancy maps and make short-term predictions into the future. To accomplish this, three different Hilbert Maps are maintained: $\mathcal{H}_p$, representing the previous timestep; $\mathcal{H}_c$, representing the current timestep; and $\mathcal{H}_a$, representing the accumulated model that is iteratively constructed as more data is collected. We start by describing how to segment objects and calculate motion between timesteps from $\mathcal{H}_p$ to $\mathcal{H}_c$, followed by the dynamic model that tracks this motion over time. We then show how to iteratively update the feature vector that defines $\mathcal{H}_a$, modifying the shape of its RKHS to account for external motion. These three steps are discussed below. 

\subsection{Object Segmentation}
\label{sec:segment}

We assume that, at each timestep $t$, a new pointcloud $\mathcal{D}_t$ is obtained, containing sensor data collected at that instant. This pointcloud is clustered to produce $\tilde{\mathcal{X}}_t=\{ \tilde{\mathcal{X}}_t^o , \tilde{\mathcal{X}}_t^f\}$, where $\tilde{\mathcal{X}}_t^o$ is the set containing clusters generated from occupied points and $\tilde{\mathcal{X}}_t^f$ contains clusters generated from unoccupied (free) points. We employ the Quick-Means algorithm \cite{GuiRam2017rss}, due to its computational speed and ability to generate similar cluster densities given a resolution threshold $r_c$. To segment individual objects $\mathcal{O}_t^p = \{ \tilde{\mathcal{X}}_t^{oq}\}_{q=1}^{Q^p}$, where $q$ are unique indexes from $\{0,...,M^o\}$, only occupied clusters in $\tilde{\mathcal{X}}_t^o$ are considered, as shown in Algorithm \ref{alg:detection}. Note that the original pointcloud $\mathcal{D}_t$ is no longer used, only the extracted clusters ${\tilde{\mathcal{X}}}_t^o$, which contributes to a much faster computational time, because $M^o \ll N$. To account for random sensor and environment artifacts, objects with fewer clusters than a certain threshold $n_c$ may be discarded.

\begin{algorithm}[!t]
	\caption{Object Segmentation Algorithm} 
	\label{alg:detection}
	\begin{algorithmic}[1]
		
		\small
		\Require 				 occupied cluster set $\tilde{\mathcal{X}}_t^o$ with $M$ entries
		\Statex  \hspace{0.34cm} cluster resolution $r_c$ 
		
		\Ensure object set $\mathcal{O}_t$ 
		
		\State $\textbf{v} \leftarrow zeros( M )$ 				  \% Object index vector
		\State $P \leftarrow 1$ 			      \hspace{1.02cm} \% Current object index
		\State $N \leftarrow \{ \textbf{n}_1 , \dots , \textbf{n}_M \} 
		\hspace{0.1cm} , \hspace{0.1cm} \textbf{n}_i \leftarrow j 
		\hspace{0.2cm} | \hspace{0.2cm} || \m_i - \m_j || < r_c $
				
		\For{ $ i \in \{1,\dots,M\}$ }
			\If{ $\textbf{v}[i] = 0$ } 				     \hspace{0.55cm} \% If not assigned
				\State $\textbf{v}[i] \leftarrow P$++    \hspace{0.8cm}  \% Start new object
				\State $recursive( N , \textbf{v} , i )$ 			 	 \% Recursive assignment
			\EndIf
		\EndFor		
		
		\State $\mathcal{O}_t \leftarrow \{\}$ \% Empty list of objects
		\For{ $i \in \{1,\dots,P\}$ }
			\State $\mathcal{O}_t^i \leftarrow \tilde{\mathcal{X}}_t^o [ \textbf{v} = i ]$ \% Add new object to list
		\EndFor
		
	\vspace{0.2cm}	
			
	\Function{Recursive}{$N,\textbf{v},i$} 		 \% Recursive function
		\For{ $ j \in N[i]$ } 				
			\State $ k \leftarrow N[i][j]$	 	\hspace{1.3cm} \% Store neighbor index
			\If{ $\textbf{v}[k] \neq \textbf{v}[i]$ } 			\hspace{0.25cm} \% If not the same object
				\State $\textbf{v}[k] \leftarrow \textbf{v}[i]$ \hspace{0.95cm} \% Assign object index
				\State $recursive( N , \textbf{v} , k )$ 					   \% Recursive assignment
			\EndIf
		\EndFor					
	\EndFunction
		
	\end{algorithmic}
\end{algorithm}

\subsection{Motion Calculation}
\label{sec:motion}

\begin{figure*}[!t]
    \vspace{-0.4cm}
	\centering
		\hspace{-0.4cm} \subfloat[Input pointcloud at timestep $t$]{\includegraphics[width=0.22\textwidth]{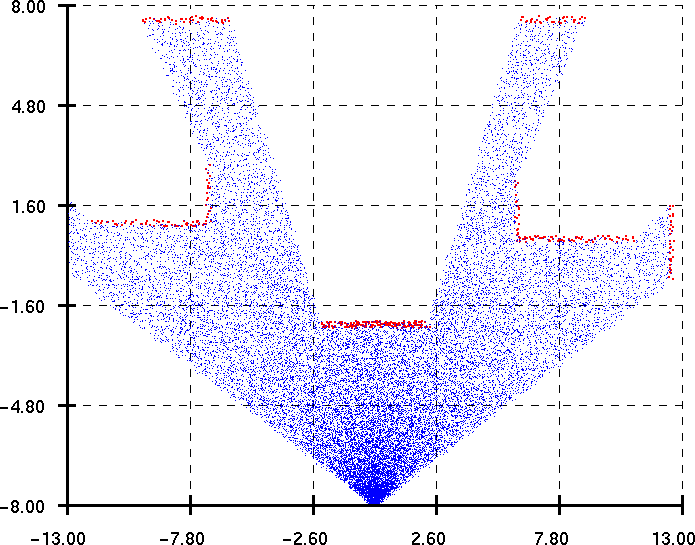}} 
        \hspace{0.1cm}
		\subfloat[Extracted clusters at timestep $t$]{	\includegraphics[width=0.22\textwidth]{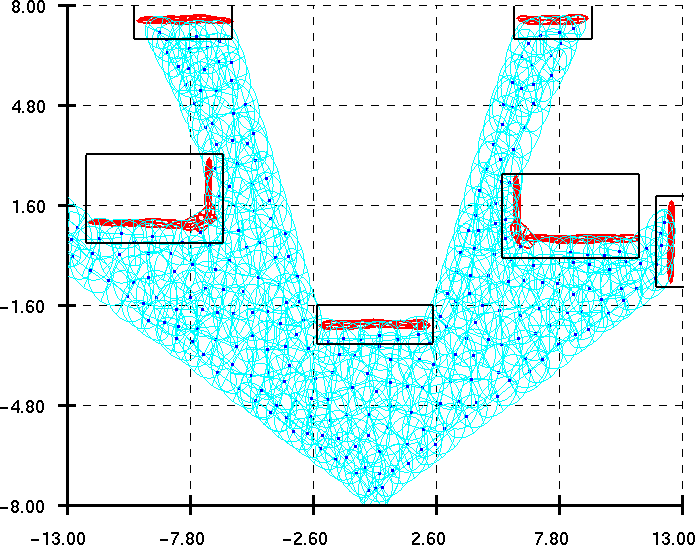}} 
        \hspace{0.1cm}
		\subfloat[Motion between $t$ and $t+1$]{ \includegraphics[width=0.22\textwidth]{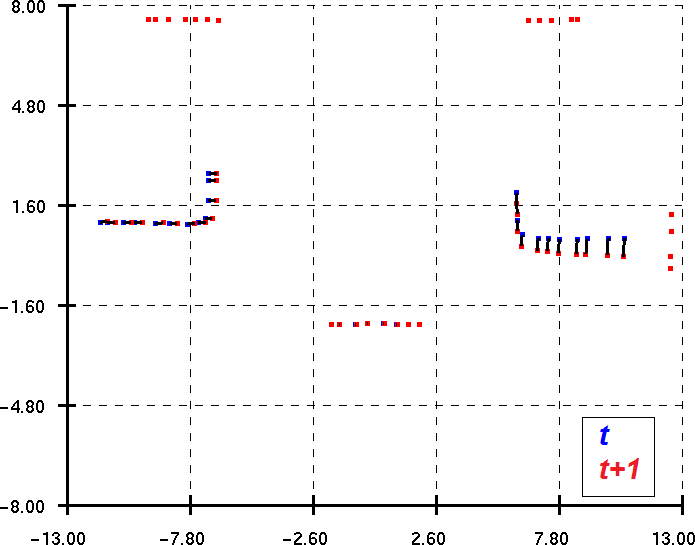}}
        \hspace{0.1cm}        
        \subfloat[Transformation]{\includegraphics[width=0.27\textwidth]{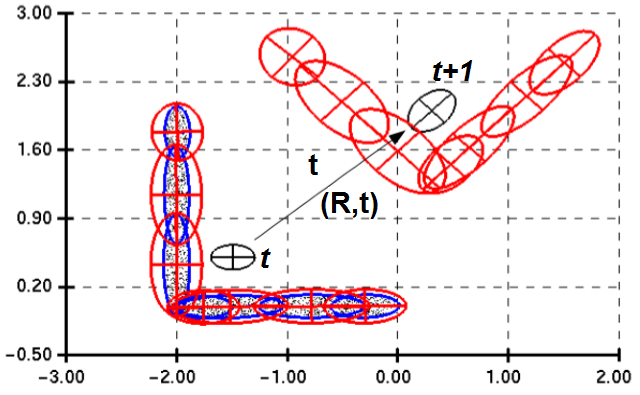}} \\
		%\hspace{-0.2cm} \subfloat[Input pointcloud at timestep $t+1$]{	\includegraphics[width=0.28\textwidth,height=3.4cm]{figures/next_pts.png}} \hspace{-0.3cm}
		%&\subfloat[Extracted clusters at timestep $t+1$]{	\includegraphics[width=0.28\textwidth,height=3.4cm]{figures/next_ctrs.png}} \hspace{-0.3cm}
	%\end{tabular}
	\caption{\small Example of object segmentation and motion calculation in simulated 2D data. (a) blue and red dots indicate unoccupied and occupied areas, respectively, (b) extracted clusters (including covariance ellipses and object boundaries). (c) Example of how location uncertainty is propagated into mapping. At $t$ the object pointcloud (black dots) is used to generate the cluster set $\tilde{\mathcal{X}}$ (blue ellipses), and the object location uncertainty $\textbf{P}$ (black ellipse) is propagated to the covariance $\bS$ of these clusters (red ellipses). At $t+1$, given the transformation ($\bR,\bt$) its location can be estimated and used to determine new cluster positions and covariances. In (d) the estimated motion is depicted, as black lines connecting their position before (blue dots) and after (red dots) the $(\bR,\textbf{t})$ ICP transformation.
}	
	\vspace{-0.4cm}
	\label{fig:motion}
\end{figure*}

Once the object set $\mathcal{O}_t = \{\mathcal{O}_t^p\}_{p=1}^{P}$ is determined, the next stage is to calculate its motion between timesteps from $t$ to $t+1$. This is done by first calculating the object set $\mathcal{O}_{t+1}$, obtained from the cluster set $\tilde{\mathcal{X}}_{t+1}^o$ extracted from $\mathcal{D}_{t+1}$. To associate between objects from different timesteps we use an overlapping metric, in which each object $\mathcal{O}_{t}^i$ is associated to the $\mathcal{O}_{t+1}^j$ with the closest cluster in its own cluster set:	
\begin{align}
\mathcal{O}_{t}^i \leftrightarrow \mathcal{O}_{t+1}^j 
\hspace{0.1cm} | \hspace{0.1cm} 
&\argmin_{j} \Big\{ \hspace{0.1cm} 
|| \m_t^{iu} - \m_{t+1}^{jv} ||_2 \hspace{0.1cm} , \nonumber  \\[-1.0ex]
& 
\m_t^{iu} \in \mathcal{O}_{t}^i 
\hspace{0.15cm} \text{and} \hspace{0.15cm}
\m_{t+1}^{jv} \in \mathcal{O}_{t+1}^j   
 \Big\},
\end{align}
where $iu$ and $jv$ are indexes representing the various clusters that belong respectively to objects $\mathcal{O}^i$ and $\mathcal{O}^j$. If the closest cluster is above a certain threshold $d_c$, that object is considered to have no association with $\mathcal{O}_{t+1}$. Similarly, objects in $\mathcal{O}_{t+1}$ that have no association are considered new (i.e. observed for the first time). Once objects $(\mathcal{O}_t^i,\mathcal{O}_{t+1}^j)$ have been associated, motion is estimated using the Iterative Closest Point (ICP) algorithm \cite{PomColSieg2015}, that minimizes the difference between two pointclouds by calculating the transformation that best matches a source to its target. Following a classical implementation \cite{SegHaeThr2009}, the two-step optimization process described below is adopted, where $\textbf{R}$ is the orthogonal transformation, $\textbf{t}$ is the translation vector and $\mathrm{SO}(D)$ is a rotation group in the $D$-dimensional Euclidean space:
\vspace{-0.1cm}
\begin{align} \small
\argmin_{ u \leftrightarrow v } &\sum_{v=1}^{Q^j} \phi( \bR \m_{t}^{iu} + \textbf{t} , \m_{t+1}^{jv} ) + I_{\mathcal{O}_{t+1}^i}(\m_{t+1}^{jv}) 
\label{eq:translation} \\
\argmin_{(\bR,\textbf{t})^{ij}_t} &\sum_{v=1}^{Q^j} \phi( \bR \m_{t}^{iu} + \textbf{t} , \m_{t+1}^{jv} ) + I_{SO(D)}(\bR),
\label{eq:rotation}
\end{align}
that alternates between computing cluster correspondences from $\mathcal{O}_{t}^i$ to $\mathcal{O}_{t+1}^j$ and solving the optimal rigid transformation $(\bR,\textbf{t})^{ij}_t$ that best aligns both sets. In the above equations, $I_A(b)$ is an indicator function that evaluates to $0$ if $b \in A$ and to $+\infty$ otherwise, and $\phi(\textbf{x},\textbf{y})$ is an error function, here selected as the Euclidean norm $||\textbf{x}-\textbf{y}||_2$. Due to a high percentage of outliers and presence of incomplete data, a sparse version of ICP \cite{BouTagPau2013} was used during experiments. An example of the proposed method is shown in Figure \ref{fig:motion} (a)-(c), where we can see the estimated cluster movement.

\subsection{Dynamic Object Model}
\label{sec:dynamic}

Once the ($\bR,\textbf{t})^{ij}_t$ transformations are obtained, they can be used to generate the dynamic models that describe object motion over time. These dynamic models allow us to: 1) estimate object position in future (or past) timesteps; 2) incorporate new observations to improve predictions; and 3) account for sensor and model uncertainties. We use a Kalman filter (KF) \cite{WelBis1995} due to its computational efficiency and closed-form parametric solution, however any other similar technique could be equally applied, such as the Gaussian process regression model from \cite{SenOttCalRam2016}.

We start by defining the $6$-dimensional state vector $\textbf{x}_t^{p} = \{ \m , \theta , \dot{\m} , \dot{\theta} \}$ for each object $\mathcal{O}_t^p$, alongside its $6 \times 6$ covariance matrix $\bP_t^{p}$ (initialized as zero). Note that this state vector includes $(x,y)$ coordinates and orientation $\theta$, which is initialized as the direction indicated by the largest eigenvector of $\bS_t^{p}$ (corresponding derivatives, i.e. velocities, are also considered). For each new timestep, the current state of objects is first propagated using the KF \textit{Prediction} equations:
\vspace{-0.4cm}
\begin{align} \small
\x_{t+1|t} &= \textbf{F}\x_{t|t}  \label{eq:prediction1} \\
\bP_{t+1|t} &= \textbf{F}\bP_{t|t} \textbf{F}^\top + \textbf{Q}. \label{eq:prediction2}
\end{align} 

\vspace{-0.1cm}
Afterwards, if a particular object is reobserved, these estimates are refined using the KF \textit{Update} equations, based on the observation $\textbf{z}_{t+1}^{p} = \{ \textbf{t}_{t+1}^{(0)p},\textbf{t}_{t+1}^{(1)p},\tan^{-1}( \bR_{t+1}^{(10)p} / \bR_{t+1}^{(00)p} )\}$ as defined in Equations \ref{eq:translation} and \ref{eq:rotation}. Note that a cluster is never truly reobserved, since it is randomly generated from pointcloud data, only its motion based on object alignment:
\begin{align} 
\x_{t+1|t+1} &= \x_{t+1|t} + \textbf{K}_{t+1}\tilde{\textbf{y}}_{t+1} \label{eq:update1} \\ 
\bP_{t+1|t+1} &= (\textbf{I}-\textbf{K}_{t+1}\textbf{H})\bP_{t+1|t}, \label{eq:update2}
\end{align}
where $\textbf{K}_{t+1} = \bP_{t+1|t}\textbf{H}^\top \textbf{S}_{t+1}^{-1}$ is the Kalman gain, with $\textbf{S}_{t+1} = \bR + \textbf{H}\bP_{t+1|t} \textbf{H}^\top$, and $\tilde{\textbf{y}}_{t+1} = \textbf{z}_{t+1} - \textbf{H} \x_{t+1|t}$ is the pre-fit measurement residual. In the above equations, $\textbf{Q}$ and $\bR$ are respectively the $6\times6$ process and $3\times3$ observation noise matrices, defined based on system configuration. During experiments, both matrices were defined using cluster resolution, such that $\textbf{Q} = \bR = r_c \cdot \textbf{I}$, to account for cluster granularity. The matrices $\textbf{F}$ and $\textbf{H}$ are respectively the state transition and observation models, defined as:
\begin{equation}
\small{
\textbf{F} = \left[ 
\begin{array}{cccccc}
1 & 0 & 0 & \Delta t & 0 & 0 \\
0 & 1 & 0 & 0 & \Delta t & 0 \\
0 & 0 & 1 & 0 & 0 & \Delta t \\
0 & 0 & 0 & 1 & 0 & 0 \\
0 & 0 & 0 & 0 & 1 & 0 \\
0 & 0 & 0 & 0 & 0 & 1 
\end{array}
\right] , \hspace{0.1cm} 
\textbf{H} = \left[ 
\begin{array}{ccc}
0 & 0 & 0 \\
0 & 0 & 0 \\
0 & 0 & 0 \\
1 & 0 & 0 \\
0 & 1 & 0 \\
0 & 0 & 1 
\end{array}
\right]^\top ,} 
\end{equation} 
where $\Delta t$ is the interval between timesteps. Even though here we focus on 2D holonomic motion, the described dynamic object model can be trivially extended to 3D navigation.

\begin{figure}[!b]
    \vspace{-0.6cm}
	\centering
    \subfloat[$t=0$]{\includegraphics[width=0.12\textwidth]{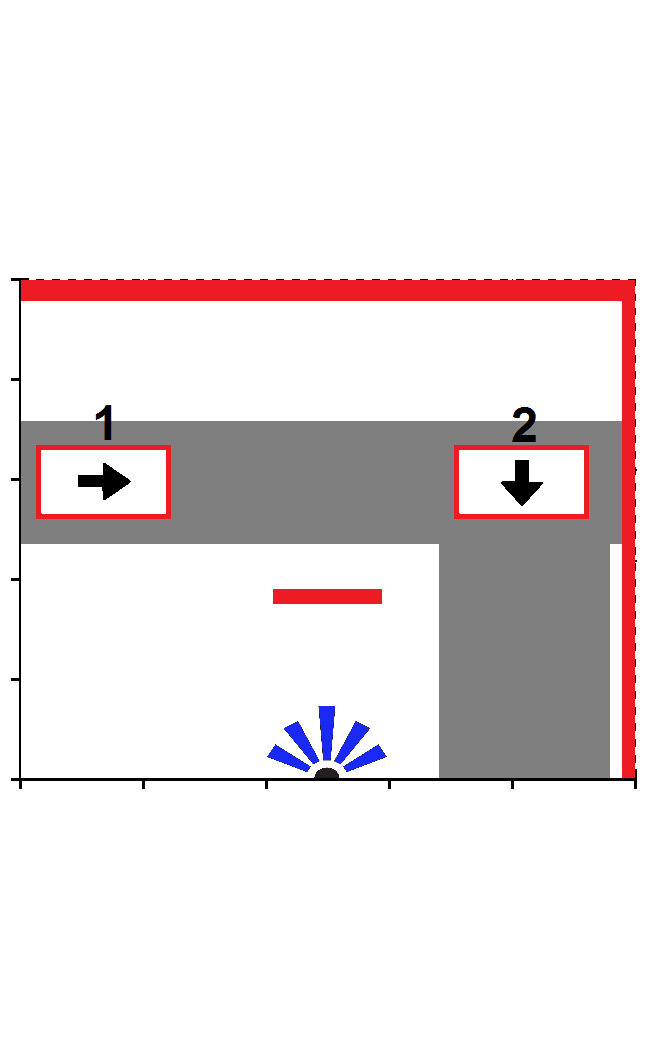}} 
	\subfloat[$t_*=0$]{\includegraphics[width=0.12\textwidth]{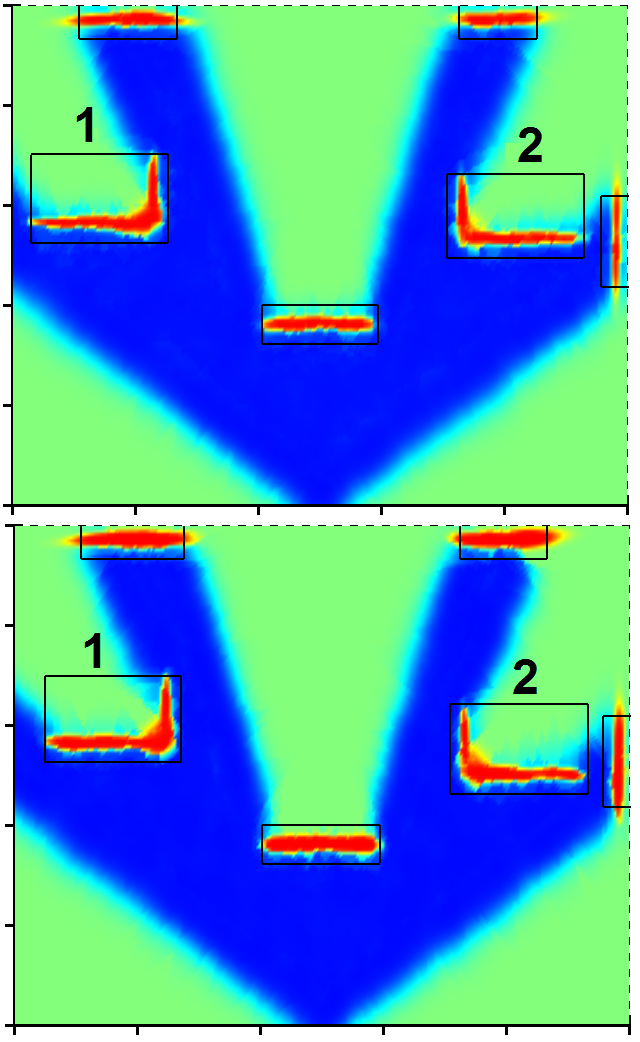}} 
	%\subfloat{\includegraphics[width=0.18\textwidth]{figures/curr_12.png}} 
	\subfloat[$t_*=18$]{\includegraphics[width=0.12\textwidth]{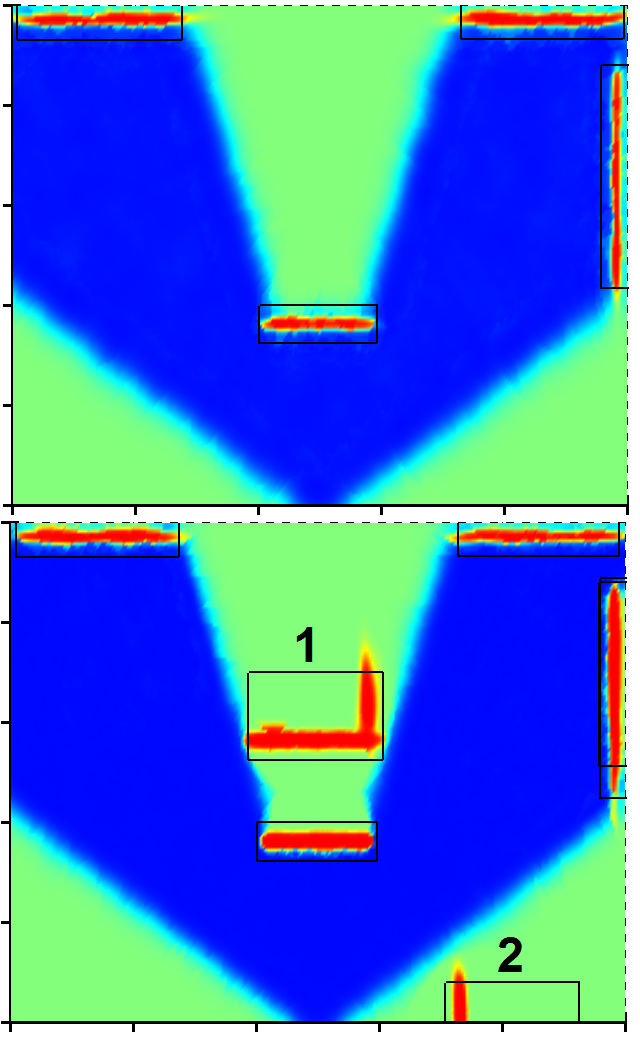}} 
	%\subfloat{\includegraphics[width=0.18\textwidth]{figures/curr_24.png}} 
	\subfloat[$t_*=42$]{\includegraphics[width=0.12\textwidth]{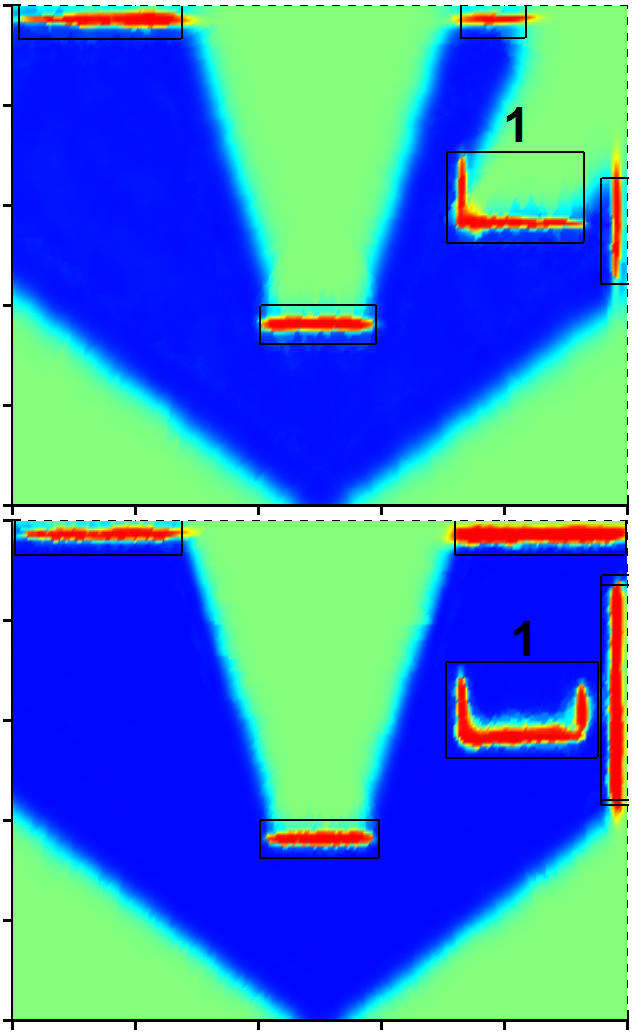}} 
	%\setcounter{subfigure}{0}			
	%\subfloat[$t=0$]{\includegraphics[width=0.12\textwidth]{figures/main_0.png}} 
	%\subfloat[$t=10$]{\includegraphics[width=0.18\textwidth]{figures/main_12.png}} 
	%\subfloat[$t=20$]{\includegraphics[width=0.12\textwidth]{figures/main_18.png}} 
	%\subfloat[$t=30$]{\includegraphics[width=0.18\textwidth]{figures/main_24.png}} 
	%\subfloat[$t=40$]{\includegraphics[width=0.12\textwidth]{figures/main_42.png}} 
	\caption{\small Example of the proposed DHM framework on simulated 2D data. (a) LiDAR (blue) observing a dynamic environment with two moving vehicles (direction in arrows) and obstacles (red). (b)-(d) Future predictions. The top row shows different $\mathcal{H}_c$, generated from current sensor data, while the bottom row shows the corresponding $\mathcal{H}_a$, generated by incremental propagation between time-steps. (b) indicates $\mathcal{H}_a$ is robust against occlusions, projecting motion into areas outside the field of view. Because of the parameter accumulation process, vehicle 1 in (c) of $\mathcal{H}_a$ has correctly mapped both sides of the vehicle.}	
	\label{fig:accexample}
\end{figure}

\subsection{Feature Vector Updates}
\label{sec:updates}

Section~\ref{sec:hm} described the feature vector $\Phi(\x,\tilde{\mathcal{X}})$ used to define the high-dimensional space in which classification takes place within the Hilbert Maps framework. Here we show how this feature vector can be iteratively updated to incorporate motion between timesteps. Instead of constantly retraining the occupancy model to account for dynamic objects, the projective function that defines the RKHS itself is modified, so the same occupancy model is able to naturally describe a changing environment and predict future states. 

At $t=0$ where there is no observed motion, an initial occupancy model $\mathcal{H}_c = \mathcal{H}_a$ is generated based on available data (see Section \ref{sec:segment}). In subsequent timesteps, the current occupancy model is reassigned as $\mathcal{H}_p$ and a new $\mathcal{H}_c$ is generated based on newly acquired data, followed by the object segmentation and alignment techniques from Section \ref{sec:segment}, relative to $\mathcal{H}_p$. Once alignment is complete, motion from associated objects is calculated according to Section \ref{sec:motion}, determining the transformations $(\bR,\bt)_t^{ij}$ that propagate different object states from $\mathcal{H}_p$ to $\mathcal{H}_c$.

These same transformations are used to propagate object states in $\mathcal{H}_a$ from $t$ to $t+1$, as shown in Section \ref{sec:dynamic}. If an object is not reobserved, only the \textit{Prediction} step is performed (Equations \ref{eq:prediction1} and \ref{eq:prediction2}), which leads to an increase in location uncertainty, otherwise the \textit{Update} step (Equations \ref{eq:update1} and \ref{eq:update2}) also takes place, which decreases location uncertainty values. These state transitions also serve to update the occupied clusters $\tilde{\mathcal{X}}_a^o=\{\bm{\mu},\bS,\omega \}_{m=1}^{M}$, as depicted in Figure~\ref{fig:motion} (d), according to the following equations:
\begin{align} \small
\m_{t+1} &= \m_{t} + \textbf{t} \\
\bS_{t+1} &= \bR_{t} \cdot \bS_{t} \cdot \bR{t}^\top \\
\omega_{t+1} &= \rho \cdot \omega_{t}. \label{eq:decay}
\end{align}

As shown above, the contribution parameter $\omega$ for each cluster is also updated, reflecting the decay caused by an increase in location uncertainty. This decay is proportional to the ratio between $||\bS||$ and $||\bS + \textbf{P}||$, i.e cluster covariance area and sum of cluster and object covariance area. Since these matrices define ellipses, this ratio can be expressed as $\rho = \nicefrac{ \prod_{i=1}^D \lambda_i^{\bS}}{ \prod_{i=1}^D \lambda_i^{\bS + \textbf{P}}}$, where $\lambda_{i}^A$ are the eigenvalues of $A$. Intuitively, $\rho=1$ if $\textbf{P} = \textbf{0}$ (no decay), and as $\textbf{P}$ increases $\rho$ decreases, which indicates a decay in occupancy confidence. Note that, during training and inference, we use $\bS_{t+1}' = \bS + \textbf{P}$ as the cluster covariance, so the probabilistic occupancy model also reflects location uncertainty.

Once this time-propagation process is complete, clusters from $\mathcal{H}_c$ are selectively incorporated into $\mathcal{H}_a$, as a way to account for newly collected data and to correct alignment errors. Clusters belonging to re-observed objects are added and receive the same dynamic model, while clusters belonging to objects observed for the first time are added with a new dynamic model, initialized to identity states. Similarly, unoccupied clusters are added without a dynamic model (i.e. are considered static). To avoid an unbound increase in cluster number and maintain density roughly constant, only clusters from $\mathcal{H}_c$ with nearest neighbor in $\mathcal{H}_a$ further than $\nicefrac{r_c}{2}$ are incorporated. The resulting cluster set defines the new feature vector $\Phi(\x,\tilde{\mathcal{X}}_a)$ that projects input data to the RKHS in which $\mathcal{H}_a$ operates. Afterwards, $\mathcal{H}_a$ is retrained using data from $\mathcal{D}_c$, to produce the current accumulated occupancy model used for inference.

\begin{table*}[!b]
	\vspace{-0.4cm}
	\begin{minipage}[b]{0.72\linewidth}
		\centering
        		\caption{\small 2D occupancy prediction results using different dynamic modeling techniques, for increasing future time steps (F-Measure scores, average of 10 runs). $HM$: Standard HM framework, without temporal modeling; $DGP$ \cite{CalRam2015}; $STHM$ \cite{SenOttCalRam2016}; Rigid Scene Flow \cite{DewCasTipBur2016}; and $DHM$: the proposed technique.}
		\begin{tabular}{|c||c|c|c|c|c||c|c|c|c|c|}
			\hline	
			{\bf Time step} & \multicolumn{5}{|c||}{\bf Dataset 1} & \multicolumn{5}{|c|}{\bf Dataset 2} \\
			\hline
			& HM & RSF & DGP & STHM & DHM 
            & HM & RSF & DGP & STHM & DHM \\
			\hline					
			$t_*=0$   & $0.824$ & $0.802$ & $0.788$ & $0.839$ & $0.844$ 
            		  & $0.807$ & $0.774$ & $0.750$ & $0.791$ & $0.811$ \\
			$t_*=1$   & $0.707$ & $0.741$ & $0.752$ & $0.784$ & $0.826$ 
            		  & $0.669$ & $0.751$ & $0.711$ & $0.765$ & $0.791$ \\
			$t_*=3$   & $0.581$ & $0.678$ & $0.678$ & $0.691$ & $0.803$  
            		  & $0.562$ & $0.668$ & $0.618$ & $0.657$ & $0.771$ \\
			$t_*=5$   & $0.418$ & $0.611$ & $0.571$ & $0.653$ & $0.756$  
            		  & $0.409$ & $0.534$ & $0.524$ & $0.607$ & $0.719$ \\
			$t_*=8$   & $0.345$ & $0.522$ & $0.502$ & $0.542$ & $0.710$ 
            		  & $0.342$ & $0.395$ & $0.414$ & $0.532$ & $0.653$ \\
			$t_*=10$  & $0.139$ & $0.478$ & $0.419$ & $0.524$ & $0.663$  
            		  & $0.104$ & $0.341$ & $0.368$ & $0.505$ & $0.619$ \\
			\hline 						
		\end{tabular}
		\label{tab:comp2D}
	\end{minipage}\hfill
	\begin{minipage}[b]{0.26\linewidth}
		\centering
		\includegraphics[width=1.0\textwidth,height=2.5cm,right]{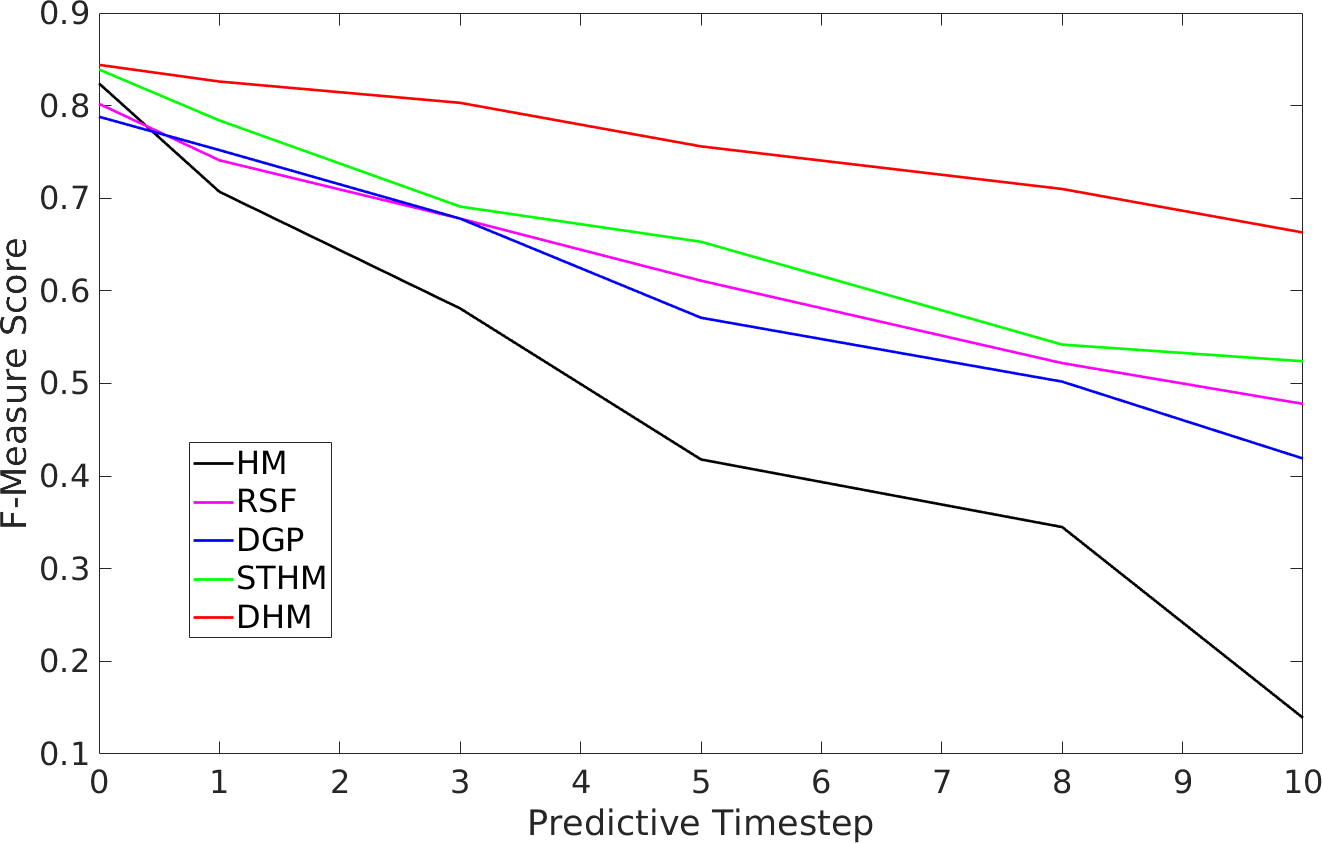}
		\captionof{figure}{{\small Plot of Table \ref{tab:comp2D}, depicting F-Measure scores for different dynamic modeling techniques, for increasing time-steps (Dataset 1).}}
		\label{fig:image}
	\end{minipage}
\end{table*}

Under the DHM framework, it is possible to query the occupancy $p(y_*=1 \vert \mathbf{x}_*, t_*,\mathbf{w},\mathcal{H}_p,\mathcal{H}_a,\mathcal{D}_{t})$ anywhere in the space $ \mathbf{x}_*$ at anytime $t_*$ (past, present, and future). Figure \ref{fig:accexample} shows an example of the proposed framework, as introduced in this section. At $t=0$ the accumulated and current models are the same, with subsequent time-steps showing how $\mathcal{H}_a$ is able to predict the occupancy state of unobserved areas, while taking into account the presence of dynamic objects. Particularly, we can see that, as the same object is observed from different perspectives, this information is incorporated into its dynamic model and used for a full reconstruction in future time-steps. It is also worth noting that static objects are tracked as well (i.e. background walls) and propagated over time, producing a seamless transition between static and dynamic states that is common in real-world applications.

\section{Experiments}
\label{sec:experiments}

A series of experiments was performed in order to demonstrate how the proposed methodology, entitled DHM (Dynamic Hilbert Maps), can be applied to the modeling of dynamic environments in both 2D and 3D scenarios. The 2D datasets considered here are the same as in \cite{SenOttCalRam2016}, consisting of laser scans collected from a busy urban intersection, covering an angular interval of $180^\circ$ and varying maximum radii (30m for the first dataset and 100m for the second one). The 3D datasets were obtained from the KITTI Vision Benchmark Suite \cite{GeiLenStiUrt2013}, collected using Velodyne sensors from both static and moving vehicles (GPS + IMU data were used to calculate and compensate ego-motion between timesteps) as they navigate through the streets of urban environments.  

\subsection{The effect of clustering}

As a baseline, initial experiments were performed for the particular case of instant predictions ($t^*=0$). In contrast to \cite{SenOttCalRam2016}, in which features are hinged randomly or in a regular grid, the proposed methodology uses clustering to produce hinge supports, according to a predetermined cluster resolution $r_c$. The effects of changing this parameter are shown in Table \ref{tab:clusters2D}, according to different error metrics obtained by randomly sampling timesteps from the first 2D dataset and using the occupancy mapping methodology described in Section \ref{sec:hm}. As expected, the number of clusters $NC$ decreases as $r_c$ increases, thus producing a more coarse model of observed structures that naturally degrades classification performance. However, a change in resolution from $0.05m$ to $5.00m$, with a corresponding increase in computational efficiency of $442\%$, contributes to a decrease of only $0.9\%$ in $AUC$ and $1.75\%$ in $ACC$, which indicates a low sensitivity to changes in scale that can be used as a trade-off between speed and accuracy. Unless noted otherwise, all further experiments use a cluster resolution value of $r_c=0.25$, which produced the smallest $NLL$ error and highest $F$-$MEAS$ score. 

\begin{table}[t]
	\centering \small
    	\caption{\small 2D occupancy modeling using DHM (average over 100 random timesteps, for $t^*=0$). $NC$: Number of clusters; $AUC$: Area Under the ROC Curve; $NLL$: Negative Log-Loss (smaller is better \cite{Bis2006}); $ACC$: Percentage of correctly predicted labels; $F$-$MEAS$: F-Measure score; and $TIME$: time in milliseconds.}
	\begin{tabular}{|c|c|c|c|c|c|c|}
		\hline
		\bf $r_c$ (m) & \bf NC & \bf AUC & \bf NLL & \bf ACC & \bf F-MEAS & \bf TIME \\
		\hline					
		$0.05$  & $844$ & $0.993$ & $0.113$ & $98.34$ & $0.809$ & $103$ \\ 							
		$0.10$  & $756$ & $0.992$ & $0.107$ & $98.35$ & $0.805$ & $ 89$ \\ 
		$0.25$  & $645$ & $0.993$ & $0.099$ & $98.26$ & $0.811$ & $ 68$ \\ 		
		$0.50$  & $592$ & $0.994$ & $0.105$ & $97.89$ & $0.784$ & $ 45$ \\ 
		$1.00$  & $480$ & $0.993$ & $0.121$ & $97.61$ & $0.776$ &  $37$ \\ 
		$2.00$  & $324$ & $0.992$ & $0.129$ & $97.33$ & $0.758$ &  $26$ \\ 
		$5.00$  & $216$ & $0.984$ & $0.151$ & $96.59$ & $0.698$ &  $19$ \\ 
		$10.00$ & $172$ & $0.965$ & $0.243$ & $93.24$ & $0.515$ &  $14$ \\ \hline 						
	\end{tabular}
	\vspace{-0.5cm}
	\label{tab:clusters2D}
\end{table}

\begin{figure}[!t]
	\centering
	\subfloat{\includegraphics[width=0.15\textwidth]{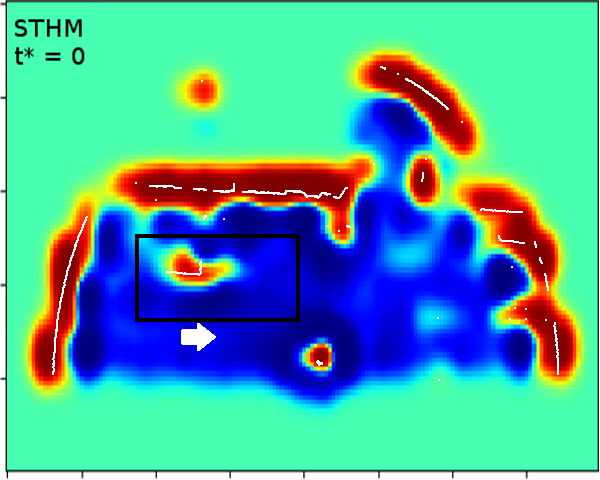}} 
	%\subfloat{\includegraphics[width=0.19\textwidth,height=2.8cm]{figures/sthm2_1.png}} 
	\subfloat{\includegraphics[width=0.15\textwidth]{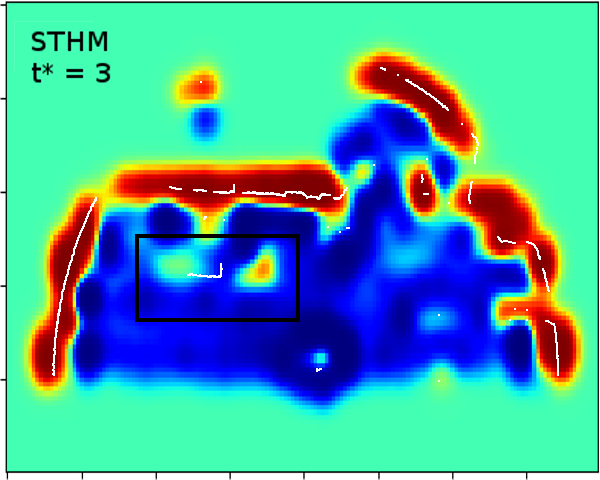}} 
	%\subfloat{\includegraphics[width=0.19\textwidth,height=2.8cm]{figures/sthm2_5.png}} 
	\subfloat{\includegraphics[width=0.15\textwidth]{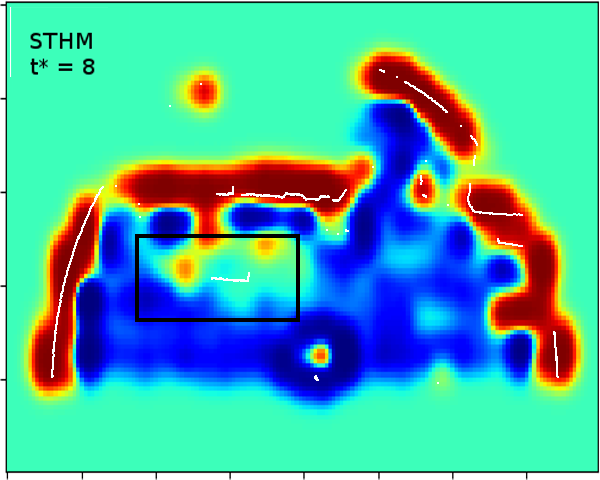}} \vspace{-0.3cm} \\	
	\subfloat{\includegraphics[width=0.15\textwidth]{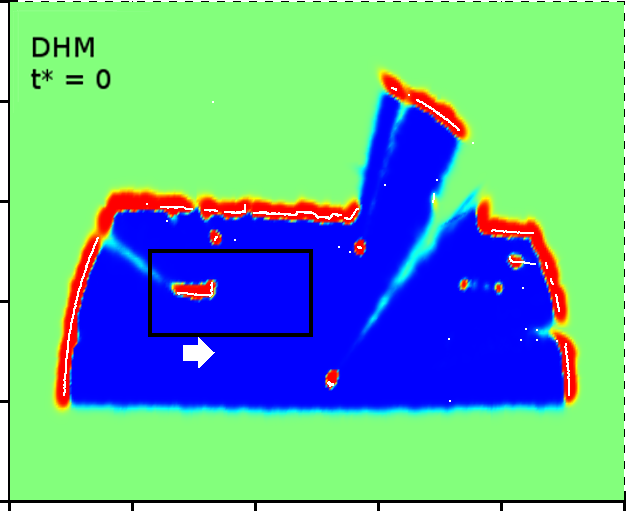}} 
	%\subfloat{\includegraphics[width=0.19\textwidth,height=2.8cm]{figures/dhm_1.png}} 
	\subfloat{\includegraphics[width=0.15\textwidth]{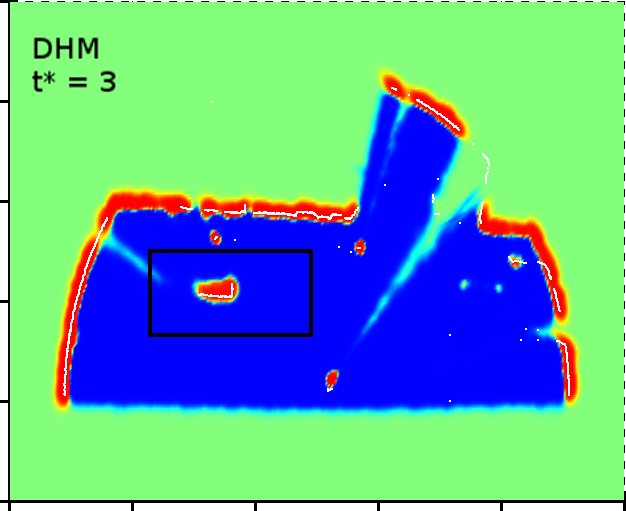}} 
	%\subfloat{\includegraphics[width=0.19\textwidth,height=2.8cm]{figures/dhm_5.png}} 
	\subfloat{\includegraphics[width=0.15\textwidth]{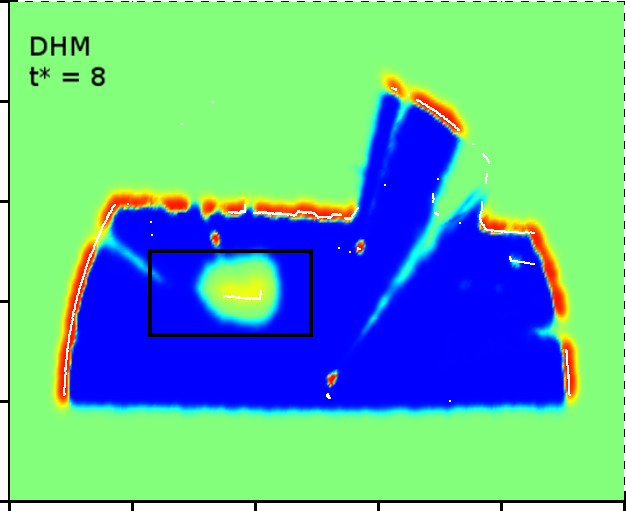}} \\ 
	\caption{\small 2D occupancy prediction results using different HM-based dynamic modeling techniques. White dots indicate current ground-truth laser reflections, and surface colors range from blue (0, unoccupied) to red (1, occupied).}	
	\label{fig:comp2d}
	\vspace{-0.5cm}
\end{figure}

\begin{figure*}[]
	\vspace{-0.7cm}
	\centering
	\subfloat[$t=0$ (pointcloud)]{\includegraphics[width=0.24\textwidth,height=3.4cm]{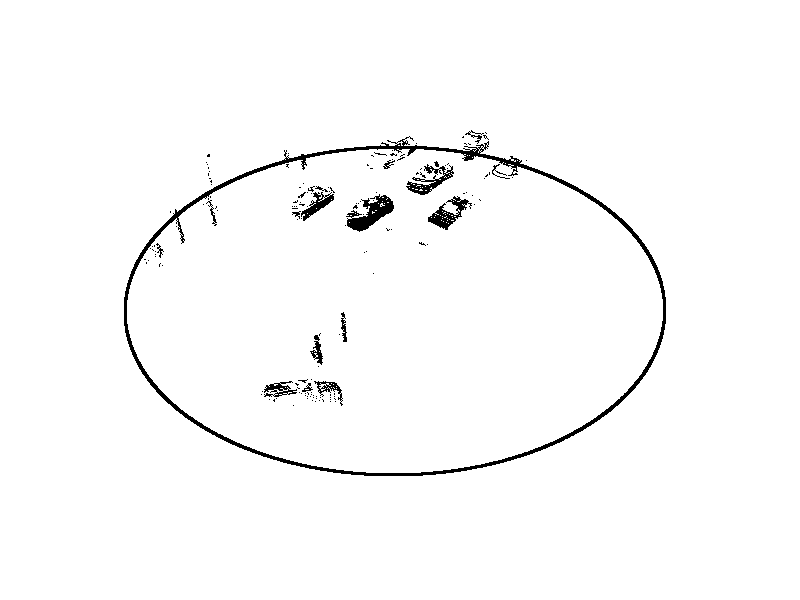}} 
	\subfloat[$t_*=0$]    		 {\includegraphics[width=0.24\textwidth,height=3.4cm]{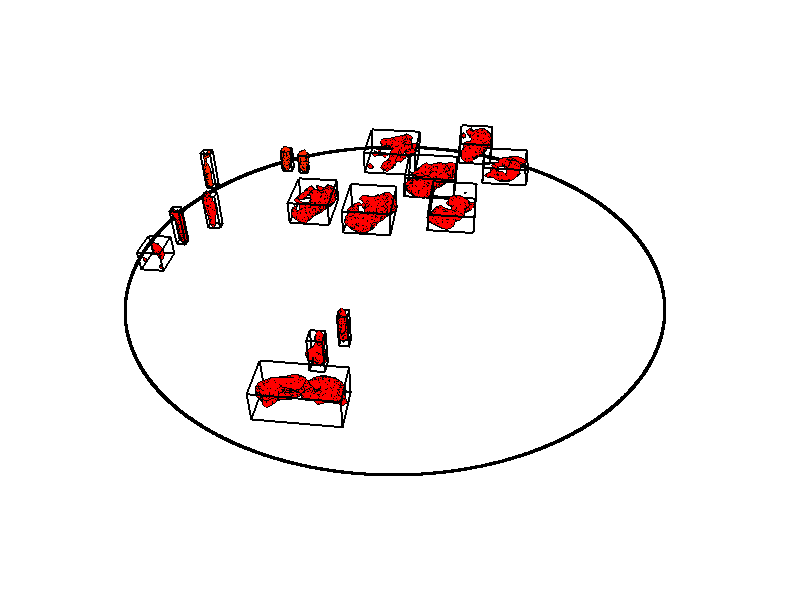}} 
	%\subfloat[$t=30$]    		 {\includegraphics[width=0.24\textwidth,height=3.4cm]{figures/tphm3d_30.png}} \\ \vspace{-0.4cm}
	\subfloat[$t_*=45$]    		 {\includegraphics[width=0.24\textwidth,height=3.4cm]{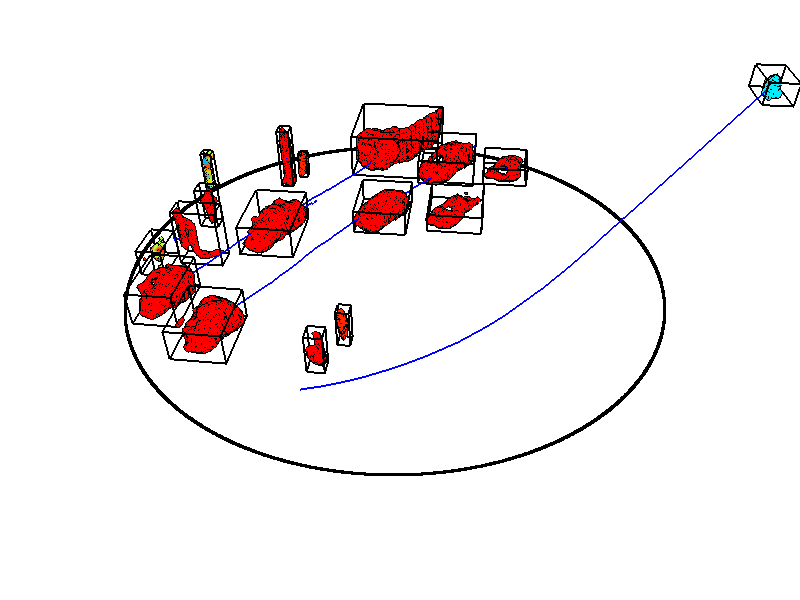}} 
	%\subfloat[$t=55$]    		 {\includegraphics[width=0.24\textwidth,height=3.4cm]{figures/tphm3d_55.png}} 
	\subfloat[$t_*=70$]    		 {\includegraphics[width=0.24\textwidth,height=3.4cm]{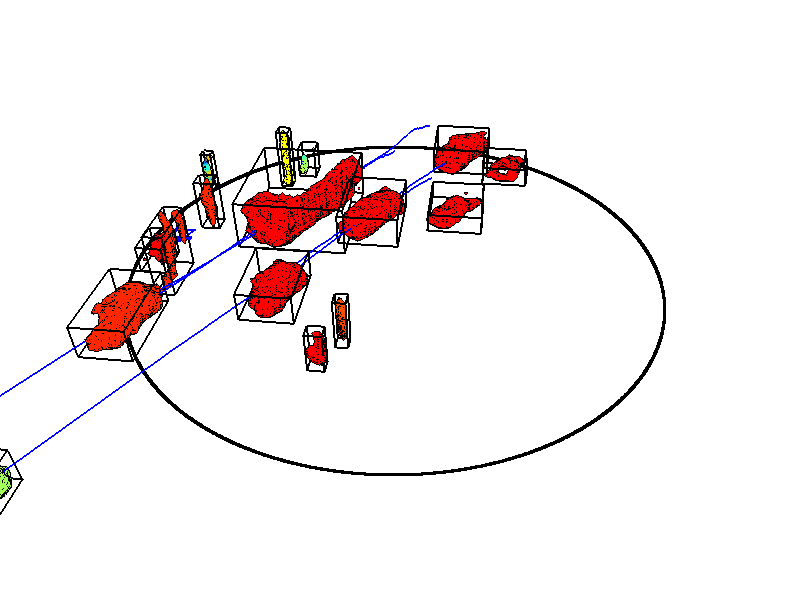}} 
	\caption{3D occupancy prediction results using the DHM framework. The black circle indicates maximum sensor range ($40m$, centered at $(0,0,0)$), and blue lines indicate object centroid motion over time. The marching cubes algorithm \cite{LorHar1987} was used for surface reconstruction, colored by weight parameter value (only clusters considered occupied, with $w >0$, are depicted). A video depicting these results can be found in \url{https://bitbucket.org/vguizilini/cvpp}.}	
	\label{fig:results3d}
	\vspace{-0.5cm}
\end{figure*}

\subsection{A comparison of dynamic occupancy maps}

A comparison between different dynamic occupancy modeling techniques, for predictions into future timesteps, is shown in Table \ref{tab:comp2D}. For all experiments, starting from an empty map, five consecutive frames are used to incrementally learn motion models for the various observed objects, and afterwards predictions are made without incorporating new information. To account for occlusions and the introduction of new objects during prediction, only areas manually annotated as containing dynamic motion (see the black rectangle in Figure~\ref{fig:comp2d}) are considered, and the F-Measure score is used due to an imbalance between classes, since it encodes both precision and recall values while being less sensitive to uncertainty increase due to weight decay. %The $RSF$ algorithm \cite{DewCasTipBur2016}, since it only segments dynamic objects and does not track them or generate occupancy maps, was used in conjunction with the dynamic model proposed in this paper, to predict the pose of dynamic objects in a deterministic way (i.e. relative to their mean center of mass).

As expected, the standard $HM$ framework, without temporal modeling, quickly degrades in performance due to un-modeled object motion, followed by $DGP$, that struggles with longer-term predictions. The two HM-based dynamic modeling techniques produce the best F-Measure scores in both datasets, however the proposed $DHM$ consistently outperforms $STHM$, while maintaining a smoother decrease in performance, mostly due to motion model inaccuracies (i.e. accumulated velocity drift errors) and the propagation of localization uncertainties to occupancy mapping estimates. The $RSF$ algorithm suffers due to its non-probabilistic nature, that is highly sensitive to small calculation errors. %It is our belief that the proposed $DHM$ framework would benefit from incorporating $RSF$ as the initial object segmentation algorithm, to improve association between pointclouds, however this is not explored here and is left for future work.   

A visual comparison between the two HM-based dynamic modeling techniques is depicted in Figure \ref{fig:comp2d}. Note that the proposed $DHM$ technique produces sharper occupancy transitions, due to the clustering process that generates non-stationary kernels for hinge support and having a clear distinction between static and dynamic objects. %As expected, both methods are able to properly track object motion over time, as highlighted in the black rectangle, however $DHM$ propagates localization mean and variance values more consistently, which we attribute to updates directly in the RKHS, rather than modifying weights belonging to fixed support points. Furthermore, since $DHM$ does not distinguish between static and dynamic objects, it propagates the state of all observed structures, as is the case of background walls that also suffer an increase in uncertainty, albeit smaller since its motion model is better defined. 

\subsection{3D dynamic maps and motion prediction}

Similar experiments were also performed using 3D datasets (Figure~ \ref{fig:results3d}), with results depicted in Table \ref{tab:comp3D}. Only $RSF$, the standard $HM$ framework and the proposed $DHM$ technique were considered, since other techniques do not scale favorably to higher dimensions ($DGP$ scales cubically with the number of training points and $STHM$ maintains a regular grid throughout the entire input space). In contrast, $DHM$ updates between timesteps require roughly $70$ ms and $120$ ms in 2D and 3D datasets\footnote{All computations were performed on a \textit{i7/2.60$\times$8 GHz} notebook, with multi-threading enabled wherever possible. A C++ demo is available at \url{https://bitbucket.org/vguizilini/cvpp}}, respectively, which makes it applicable to online tasks under real-time constraints. Interestingly, the proposed technique achieved better overall results for longer-term predictions when using 3D data, most likely due to a richer pointcloud representation of structures, that facilitates ICP alignment and thus produces better motion models. Additionally, it is worth noting that the introduction of a moving sensor for data collection did not significantly impact performance.

%A visual representation of the proposed technique is shown in Figure \ref{fig:results3d}, with a different setup, in which the occupancy states of areas beyond sensor range (black circle) are predicted according to motion models learned while objects were within range. As we can see, new vehicles are seamlessly incorporated into the framework, and as they leave their occupancy states are still propagated over time, with decaying weights representing uncertainty increase (objects are removed once weight values drop below a certain threshold). Note that static objects (i.e. poles) are also propagated over time as they become occluded by vehicles, and recovered once visible again. 

\section{Conclusion}
\label{sec:conclusion}

This paper introduces a novel technique for dynamic occupancy mapping that efficiently incorporates temporal dependencies between data collected in different time-steps. Under this framework, new observations are used to learn a global accumulation map. This enables seamlessly generating future occupancy maps even in the presence of occlusions. Considering both runtime and accuracy, the proposed framework outperforms state-of-the-art dynamic mapping techniques as tested using 2D and 3D datasets. Future work will focus on different motion models for tracking more complex patterns and improving data association between objects. 

\begin{table}[!t]
	\centering 
    	\caption{\small 3D occupancy prediction results (F-Measure).}
	\begin{tabular}{|c||c|c|c||c|c|c|}
		\hline	
		\bf Time step & \multicolumn{3}{|c||}{\bf Static Sensor} & \multicolumn{3}{|c|}{\bf Moving Sensor} \\
		\hline
		$r_c$ (m) & HM & RSF & DHM & HM & RSF & DHM\\
		\hline					
		$t_*=0$   & $0.841$ & $0.869$ & $0.858$ & $0.843$ & $0.845$ & $0.837$ \\
		$t_*=1$   & $0.729$ & $0.781$ & $0.823$ & $0.727$ & $0.758$ & $0.804$ \\
		$t_*=3$   & $0.550$ & $0.694$ & $0.776$ & $0.529$ & $0.678$ & $0.742$ \\
		$t_*=5$   & $0.459$ & $0.621$ & $0.714$ & $0.472$ & $0.612$ & $0.718$ \\
		$t_*=8$   & $0.320$ & $0.588$ & $0.670$ & $0.301$ & $0.559$ & $0.660$ \\
		$t_*=10$  & $0.168$ & $0.529$ & $0.649$ & $0.189$ & $0.498$ & $0.625$ \\
		\hline 						
	\end{tabular}
	\label{tab:comp3D}
	\vspace{-0.6cm}
\end{table}

%This paper introduces a novel technique for occupancy mapping that incorporates temporal dependencies between data collected in different timesteps. The Hilbert Maps framework is used to produce an occupancy model of the observed environment, and different objects are tracked over time to produce probabilistic motion models, that are used to propagate occupancy states into the future (or past). Different from other techniques, it performs this propagation by directly updating the RKHS in which classification takes place, both in terms of mean (i.e. object translation) and variance (i.e. localization uncertainty). New objects can be seamlessly incorporated into the framework as they are first observed, and occluded objects can be reincorporated as they become visible again. The proposed Time-Propagating Hilbert Maps framework was tested in both 2D and 3D datasets with results that outperform current state-of-the-art dynamic modeling techniques, and its computational efficiency allows deployment in online applications. Future work will focus on different motion models, for the tracking of more complex patterns, and improving data association between objects, for a better understanding of the scene and how new data is incorporated into the occupancy model. 

%%%%%%%%%%%%%%%%%%%%%%%%%%%%%%%%%%%%%%%%%%%%%%%%%%%%%%%%%%%%%%%%%%%%%%%%%%%%%%%%

\section*{ACKNOWLEDGMENT}

This research was supported by funding from the Faculty of Engineering \& Information Technologies, The University of Sydney, under the Faculty Research Cluster Program.

%%%%%%%%%%%%%%%%%%%%%%%%%%%%%%%%%%%%%%%%%%%%%%%%%%%%%%%%%%%%%%%%%%%%%%%%%%%%%%%%
%\newpage
\bibliographystyle{IEEEtran}
\bibliography{bibVitor.bib}

%%%%%%%%%%%%%%%%%%%%%%%%%%%%%%%%%%%%%%%%%%%%%%%%%%%%%%%%%%%%%%%%%%%%%%%%%%%%%%%%

\end{document}